\documentclass[conference]{IEEEtran}
\IEEEoverridecommandlockouts
\usepackage{cite}
\usepackage{amsmath,amssymb,amsfonts}
\usepackage{graphicx}
\usepackage{textcomp}
\usepackage{xcolor}
\usepackage{algorithm}
\usepackage{subcaption}
\usepackage{algpseudocode}
\usepackage{booktabs}    % For \toprule, \midrule, \bottomrule
\usepackage{multirow}    % For \multirow{}
\usepackage{rotating}    % For \rotatebox

\def\BibTeX{{\rm B\kern-.05em{\sc i\kern-.025em b}\kern-.08em
    T\kern-.1667em\lower.7ex\hbox{E}\kern-.125emX}}
\begin{document}

\title{Performance Comparison of Classical and Neural Sampling Algorithms for Robotic Navigation\\

}

\author{\IEEEauthorblockN{1\textsuperscript{st} Hichem Cheriet}
\IEEEauthorblockA{\textit{dept. of Computer Science} \\
\textit{Université d'USTO Mohamed Boudiaf}\\
Oran, Algeria \\
hichem.cheriet@univ-usto.dz}
\and
\IEEEauthorblockN{2\textsuperscript{nd} Khellat Kihel Badra}
\IEEEauthorblockA{\textit{dept. of Economics} \\
\textit{Oran2 Mohamed BenAhmed University}\\
Oran, Algeria \\
khellat\_badra@yahoo.fr}
\and
\IEEEauthorblockN{3\textsuperscript{rd} Chouraqui Samira}
\IEEEauthorblockA{\textit{dept. of Computer Science} \\
\textit{Université d'USTO Mohamed Boudiaf}\\
Oran, Algeria \\
samirachouraqui178@gmail.com}

}

\maketitle

\begin{abstract}
Integrating artificial intelligence (AI) into sampling-based motion planning opens new opportunities for more efficient autonomous navigation. In this study, three algorithms namely RRT*, Neural RRT*, and Neural Informed RRT* are implemented and compared on maps containing convex and concave obstacles with varying densities. The results show that neural-guided planners significantly improve path quality, achieving up to 14\% shorter paths and 55–75\% smoother trajectories compared to classical RRT*. Neural Informed RRT* consistently outperforms both RRT* and Neural RRT* in terms of path length and smoothness, demonstrating the potential of AI-based guidance for enhancing reliability and trajectory efficiency in robotic and UAV navigation, although planning time may increase slightly. These findings highlight the important role of AI in advancing real-time robotic path planning.
\end{abstract}


\section{Introduction}
Path planning operations are a key component of autonomous systems such as ground robotics, unmanned aerial vehicles (UAVs), underwater robots, and self-driving vehicles. Among the existing approaches, the Rapidly-Exploring Random Tree (RRT) \cite{[b0]} and its optimal variant, RRT* \cite{[b1]}, have gained wide popularity due to their efficiency, especially in high-dimensional spaces. However, these algorithms suffer from inefficient random sampling, leading to high computational cost, slow convergence, and suboptimal exploration, particularly in cluttered environments.

Several improvements have been proposed to address these limitations. For example, Informed RRT* focuses the search on a reduced ellipsoidal region, while other variants introduce heuristic strategies to accelerate convergence. Although these methods improve performance, they still rely on hand-crafted heuristics and do not adapt efficiently to different environment structures.

Recent advances in machine learning have motivated the integration of neural networks into planning frameworks to bias sampling toward promising regions. Such hybrid methods aim to combine the optimality guarantees of RRT* with the learning ability of neural models. In particular, Neural RRT* and Neural Informed RRT* introduce learned sampling distributions to guide exploration and improve path quality.
\begin{figure}[h]
    \centering
    \includegraphics[width=0.9\linewidth]{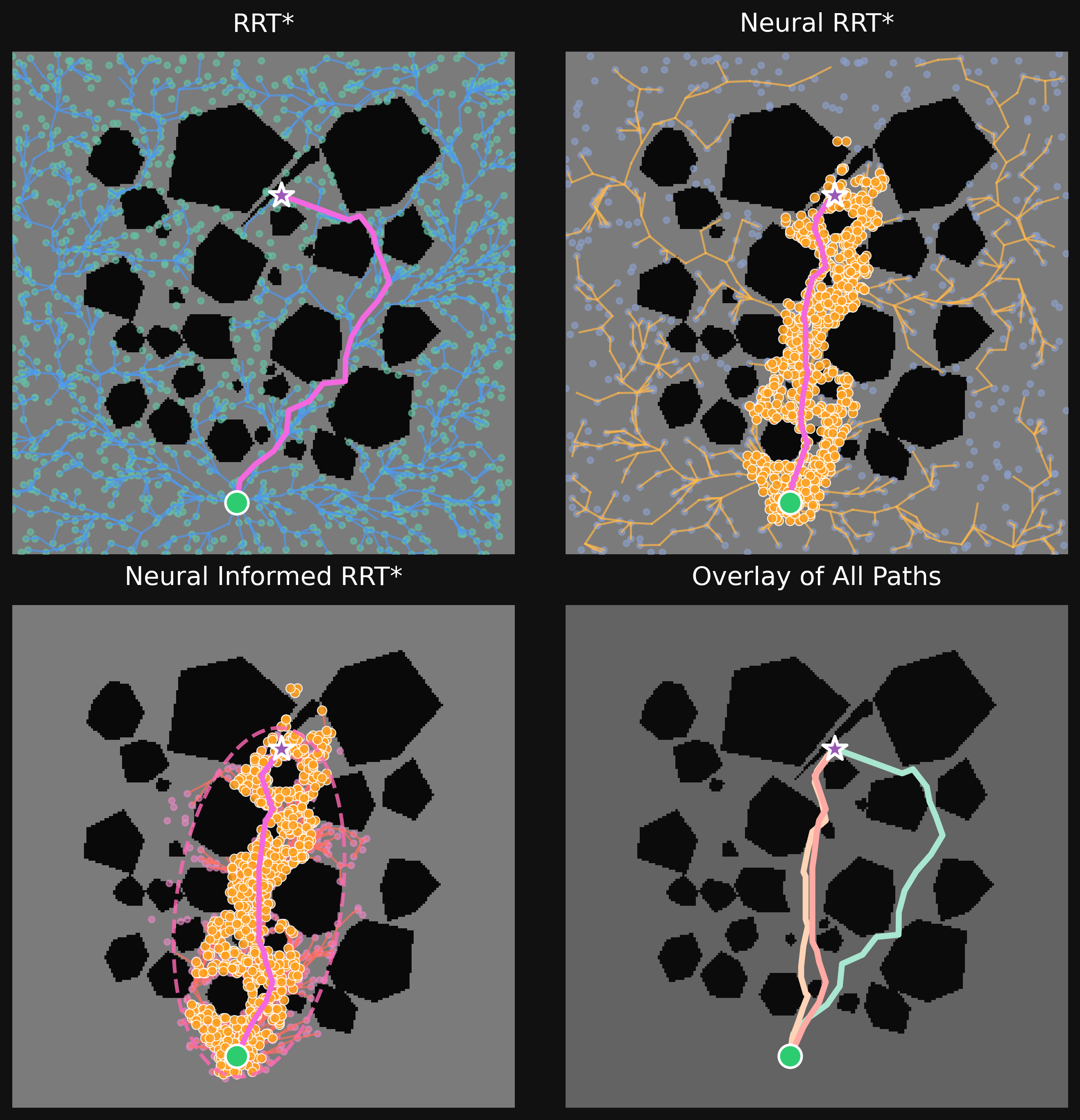}
\caption{Comparison of RRT variants for path planning. Neural guidance (orange points) and the informed sampling ellipse (dashed magenta) improve exploration and path optimality. All final paths are shown in the bottom-right.}

    \label{fig:placeholder}
\end{figure}
However, despite these advances, a clear and unified evaluation of neural-guided planners compared to classical methods 

under consistent conditions remains limited. In many existing works, algorithms are evaluated independently or under different experimental setups, making it difficult to quantify the true benefits of neural guidance in terms of efficiency, path quality, and reliability.

This paper compares the classical RRT*, a Neural RRT* \cite{[b9]}, and a Neural Informed RRT* \cite{[b10]} that incorporates learned heuristics into the informed sampling process. The comparison is conducted under a unified framework across environments with varying obstacle densities, allowing a fair assessment of the trade-offs between classical and learning-based approaches.

Fig.~\ref{fig:placeholder} provides a graphical overview of the proposed comparison, highlighting how neural guidance and informed sampling improve exploration and path optimality compared to classical RRT*.

Unlike prior works that evaluate neural planners independently, this study provides a unified and controlled comparison across multiple environments and metrics, highlighting the practical trade-offs between classical and learning-based sampling strategies.

The main contributions are:

\begin{itemize}
    \item An implementation and comparison of two AI-enhanced planning algorithms with a classical baseline on maps of varying obstacle complexity.
    \item Quantitative evaluation using planning time, path length, path smoothness, and success rate.
    \item Discussion of AI’s impact on convergence efficiency and solution quality.
\end{itemize}

\section{Related Works}

The Rapidly-exploring Random Tree (RRT) algorithm, introduced by LaValle et al. \cite{[b0]}, has become a cornerstone in sampling-based motion planning. It incrementally constructs a tree by randomly sampling the configuration space and connecting feasible nodes, which provides fast exploration even in high-dimensional environments. However, the standard RRT is not asymptotically optimal and often produces suboptimal, jagged paths that require post-processing for smoothing.

To address these limitations, Karaman et al. \cite{[b1]} proposed an enhanced variant, namely RRT*. After finding the first path, the algorithm rewires nearby nodes during the tree expansion to ensure optimality. Despite its theoretical guarantees, RRT* also suffers from high computational cost, especially in complex or cluttered spaces, where convergence toward the optimal solution can be slow and requires a lot of iterations.

Other studies have focused on accelerating convergence and improving search efficiency. The Informed RRT* \cite{[b2]} introduced an ellipsoidal sampling domain that restricts exploration to a cost-bounded region, while Smart-RRT* \cite{[b3]} enhanced path optimization through intelligent rewiring and node selection strategies. Other variants, such as Anytime RRT* \cite{[b4]}, trade optimality for faster initial solutions by iteratively refining paths as computation time allows.

Hybrid approaches combine RRT* with metaheuristic techniques to guide exploration. PSO-RRT* \cite{[b5]} and APF-RRT* \cite{[b5]} leverage particle swarm optimization and artificial potential fields, respectively, to bias sampling toward promising regions. These methods yield smoother and shorter paths, but at the expense of increased computational overhead and parameter sensitivity.

With the advancement of deep learning, neural-guided motion planning has emerged as a promising technique for enhancing sampling efficiency and decision-making in complex environments. Learning-based methods such as L2RRT \cite{[b7]}, and Motion Planning Network (MPNet) \cite{[b8]} utilize neural networks to learn sampling distributions or predict collision-free regions, which improves planning efficiency by encoding prior environmental knowledge. 

Motivated by these limitations of the classical planners, the present work provides a unified experimental comparison of RRT*, Neural RRT*, and Neural Informed RRT*, aiming to quantify the actual benefits of neural guidance in classical motion planning methods.

\section{Comparison Algorithms Overview}

This section presents the three planners evaluated in this study: RRT*, Neural RRT*, and Neural Informed RRT*. All belong to the RRT family, which is designed for efficient exploration of different environments.

\subsection{RRT*}
\begin{figure}[h]
    \centering
    \includegraphics[width=0.45\textwidth]{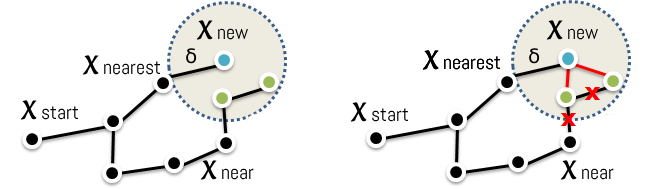}
    \caption{Illustration of the RRT* expansion process showing random samples, nearest nodes, steering, and rewiring.}
    \label{fig:rrtstar}
\end{figure}

The classical RRT* algorithm incrementally constructs a tree rooted at \( x_{\text{start}} \).  
At each iteration, a random point \( x_{\text{rand}} \) is sampled uniformly from the free configuration space \( \mathcal{X}_{\text{free}} \), denoted as \( x_{\text{rand}} \sim U(\mathcal{X}_{\text{free}}) \).  
The nearest node \( x_{\text{nearest}} = \text{Nearest}(T, x_{\text{rand}}) \) is identified, and a new node is generated by extending a step of fixed length \( \delta \) toward \( x_{\text{rand}} \):
\[
x_{\text{new}} = \text{Steer}(x_{\text{nearest}}, x_{\text{rand}}, \delta).
\]
If the local path between \( x_{\text{nearest}} \) and \( x_{\text{new}} \) is collision-free, the new node is added to the tree; otherwise, the sampling process repeats.  
A rewiring step then updates parent connections within a neighborhood radius \( r \) to minimize the path cost, ensuring asymptotic optimality (Fig.\ref{fig:rrtstar}).  
The steps of the RRT* algorithm are summarized in Alg.~\ref{alg:rrtstar}.

\begin{algorithm}[h]
\caption{RRT* Algorithm}
\label{alg:rrtstar}
\begin{algorithmic}[1]
\State Initialize tree $\mathcal{T} \leftarrow \{x_{\text{start}}\}$
\While{termination not met}
    \State Sample $x_{\text{rand}} \sim U(\mathcal{X}_{\text{free}})$
    \State $x_{\text{nearest}} \leftarrow \text{Nearest}(\mathcal{T}, x_{\text{rand}})$
    \State $x_{\text{new}} \leftarrow \text{Steer}(x_{\text{nearest}}, x_{\text{rand}}, \delta)$
    \If{CollisionFree($x_{\text{nearest}}, x_{\text{new}}$)}
        \State Add $x_{\text{new}}$ to $\mathcal{T}$ and rewire within radius $r$
    \EndIf
\EndWhile
\State \Return $\mathcal{T}$ and optimal path
\end{algorithmic}
\end{algorithm}

Although RRT* guarantees asymptotic optimality, it often converges slowly due to its uniform random sampling, particularly in cluttered environments.

\subsection{Neural RRT*}

The Neural RRT* algorithm enhances the sampling step of RRT* using a learned distribution obtained from a neural network. 
Instead of uniform sampling, random nodes are selected according to a probability map $P(x,y)$ predicted by the trained network:
\begin{equation}
x_{\text{rand}} \sim \alpha P(x,y) + (1-\alpha)U(\mathcal{X}_{\text{free}}),
\end{equation}
where $\alpha \in [0,1]$ controls the balance between learned and uniform sampling.

\begin{algorithm}[h]
\caption{Neural RRT* Algorithm}
\begin{algorithmic}[1]
\State Initialize tree $\mathcal{T} \leftarrow \{x_{\text{start}}\}$
\While{termination not reached}
    \State Predict $P(x,y)$ from CNN given current map
    \State Sample $x_{\text{rand}} \sim \alpha P(x,y) + (1-\alpha)U(\mathcal{X}_{\text{free}})$
    \State $x_{\text{near}} \leftarrow \text{Nearest}(\mathcal{T}, x_{\text{rand}})$
    \State $x_{\text{new}} \leftarrow \text{Steer}(x_{\text{near}}, x_{\text{rand}}, \eta)$
    \If{CollisionFree($x_{\text{near}}, x_{\text{new}}$)}
        \State $\mathcal{T} \leftarrow \mathcal{T} \cup \{x_{\text{new}}\}$
        \State RewireNeighbors($\mathcal{T}, x_{\text{new}}, r$)
    \EndIf
\EndWhile
\end{algorithmic}
\end{algorithm}

\subsection{Neural Informed RRT*}

The Neural Informed RRT* combines the advantages of Neural RRT* and Informed RRT*. After finding an initial feasible path with cost $c_{\text{best}}$, the planner limits sampling to an ellipsoidal region:
\begin{equation}
\mathcal{E} = \{x \in \mathbb{R}^2 \mid \|x - x_{\text{center}}\|_{C^{-1}} \le 1\},
\end{equation}
defined by $x_{\text{start}}$, $x_{\text{goal}}$, and $c_{\text{best}}$. 
Samples are selected from the intersection of this region and the neural probability map:
\begin{equation}
x_{\text{rand}} \sim P(x,y) \cap \mathcal{E}.
\end{equation}
This dual constraint focuses exploration on areas that are both geometrically promising and dynamically feasible, which improves convergence speed and path quality.

\begin{algorithm}[h]
\caption{Neural Informed RRT* Algorithm}
\begin{algorithmic}[1]
\State Initialize tree $\mathcal{T} \leftarrow \{x_{\text{start}}\}$
\State Initialize $c_{\text{best}} \leftarrow \infty$
\While{termination not reached}
    \If{$c_{\text{best}} = \infty$}
        \State Sample $x_{\text{rand}} \sim P(x,y)$
    \Else
        \State Define ellipsoid $\mathcal{E}(x_{\text{start}},x_{\text{goal}},c_{\text{best}})$
        \State Sample $x_{\text{rand}} \sim P(x,y) \cap \mathcal{E}$
    \EndIf
    \State $x_{\text{near}} \leftarrow \text{Nearest}(\mathcal{T}, x_{\text{rand}})$
    \State $x_{\text{new}} \leftarrow \text{Steer}(x_{\text{near}}, x_{\text{rand}}, \eta)$
    \If{CollisionFree($x_{\text{near}}, x_{\text{new}}$)}
        \State $\mathcal{T} \leftarrow \mathcal{T} \cup \{x_{\text{new}}\}$
        \State RewireNeighbors($\mathcal{T}, x_{\text{new}}, r$)
        \If{ReachedGoal($x_{\text{new}}$)}
            \State Update $c_{\text{best}}$
        \EndIf
    \EndIf
\EndWhile
\end{algorithmic}
\end{algorithm}

\subsection{Neural Network Model}

In this paper, both neural planners employ the same U-Net-based model to generate a sampling prior distribution \(P(x, y)\). 
The input is a three-channel $224 \times 224$ image that encodes the environment: the obstacle map, and the start and goal positions marked in red and green, respectively, providing the model with spatial context for navigation. 

The encoder–decoder structure leverages the representational power of ResNet-50 for robust spatial feature extraction, while the U-Net skip connections preserve fine-grained geometric details critical for accurate path localization.
The network outputs a dense probability map $P(x,y)$ highlighting traversable and goal-oriented regions to guide the sampling process.

\textbf{Dataset generation:} The training dataset is generated using the A* algorithm on randomly generated grid maps with varying obstacle densities (sparse, medium, and dense). A total of \textbf{N} maps are created, each with randomly sampled start and goal positions. For each map, the optimal path is computed and converted into a binary mask. To improve learning robustness, the path is dilated with a radius of 3 pixels, producing smoother supervision signals.

\textbf{Training process:} The model is trained in a supervised manner using the generated dataset. The training set consists of \textbf{X} samples, while \textbf{Y} samples are reserved for validation. The network is trained for \textbf{E} epochs using the Adam optimizer with a learning rate of \textbf{LR} and a batch size of \textbf{B}. Data augmentation techniques such as random obstacle variations and start–goal repositioning are applied to improve generalization.

The model is optimized using the Dice loss:
\begin{equation}
\mathcal{L}_{\text{Dice}} = 1 - \frac{2|P \cap Y|}{|P| + |Y|},
\end{equation}
where $Y$ denotes the ground-truth path mask obtained from A*.

\textbf{Inference:} During inference, the model predicts $P(x,y)$ in real time ($<10$ ms per map), enabling efficient online sampling guidance for both Neural RRT* and Neural Informed RRT*.

\textbf{Generalization:} The trained model is evaluated on unseen maps with different obstacle configurations to ensure that the learned sampling strategy generalizes beyond the training distribution.

\section{Results and Discussion}

\subsection{Environmental Settings}
To ensure a fair and comprehensive comparison, the planners were tested on three categories of maps with varying obstacle densities:
\begin{itemize}
    \item \textit{Sparse maps:} environments with few obstacles and wide open spaces.
    \item \textit{Medium-density maps:} environments with a moderate number of obstacles distributed across the grid.
    \item \textit{Dense maps:} environments containing many closely spaced obstacles, which represent challenging navigation scenarios.
\end{itemize}

Five random maps were generated per environment category, and each planner was executed ten times per map, yielding 50 runs per map type. All planners used 1000 iterations, a step size of 10, and a rewiring radius of 10. The sampling balance parameter was set to \(\alpha = 0.5\), following the setting used in Neural Informed RRT* \cite{[b10]}. This value provides a balanced trade-off between exploitation of the learned sampling distribution and exploration of the configuration space, ensuring robustness while maintaining efficient convergence. Experiments were conducted on the Kaggle platform using a CPU-based environment with an Intel Xeon processor running at 2.20 GHz and 30 GB RAM. Mean and standard deviation were reported for all metrics.

\subsection{Evaluation Metrics}
The performance of the three planners, RRT*, Neural RRT*, and Neural Informed RRT*, was evaluated on two-dimensional grid environments using four quantitative metrics: path length, planning time, smoothness, and success rate. These metrics were chosen to assess both the efficiency and the quality of the generated paths.

\textbf{Path length} measures the total travel distance between the start and goal points along the planned trajectory. It reflects the optimality of the path, where shorter lengths correspond to more efficient exploration and fewer unnecessary deviations.

\textbf{Planning time} represents the total computation time required to generate a feasible solution. This metric is critical for real-time and onboard applications, where rapid path generation is essential.

\textbf{Smoothness} quantifies the curvature and continuity of the path by evaluating the number and sharpness of direction changes. A smoother path indicates better dynamic feasibility and facilitates easier trajectory tracking by robots.

\textbf{Success rate} measures the percentage of successful runs in which the planner reached the goal without collisions. It serves as an indicator of the planner's reliability and robustness under different environmental conditions.

\subsection{Comparison Results}
\begin{figure}[htbp]
    \centering
    \begin{subfigure}[b]{0.7\linewidth}
        \centering
        \includegraphics[width=\linewidth]{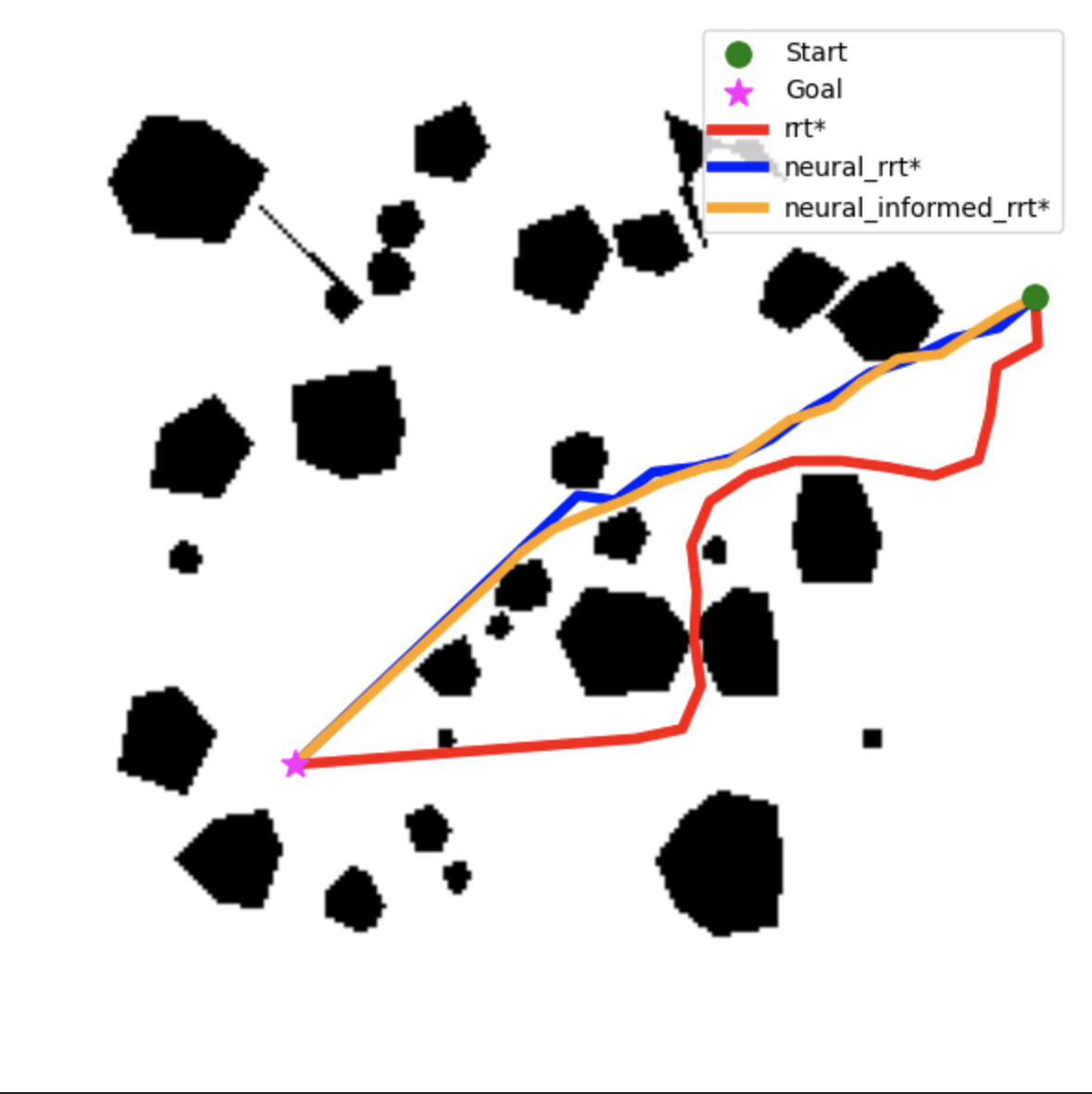}
        \caption{Sparse maps}
        \label{fig:sparse_comparison}
    \end{subfigure}
    \hfill
    \begin{subfigure}[b]{0.7\linewidth}
        \centering
        \includegraphics[width=\linewidth]{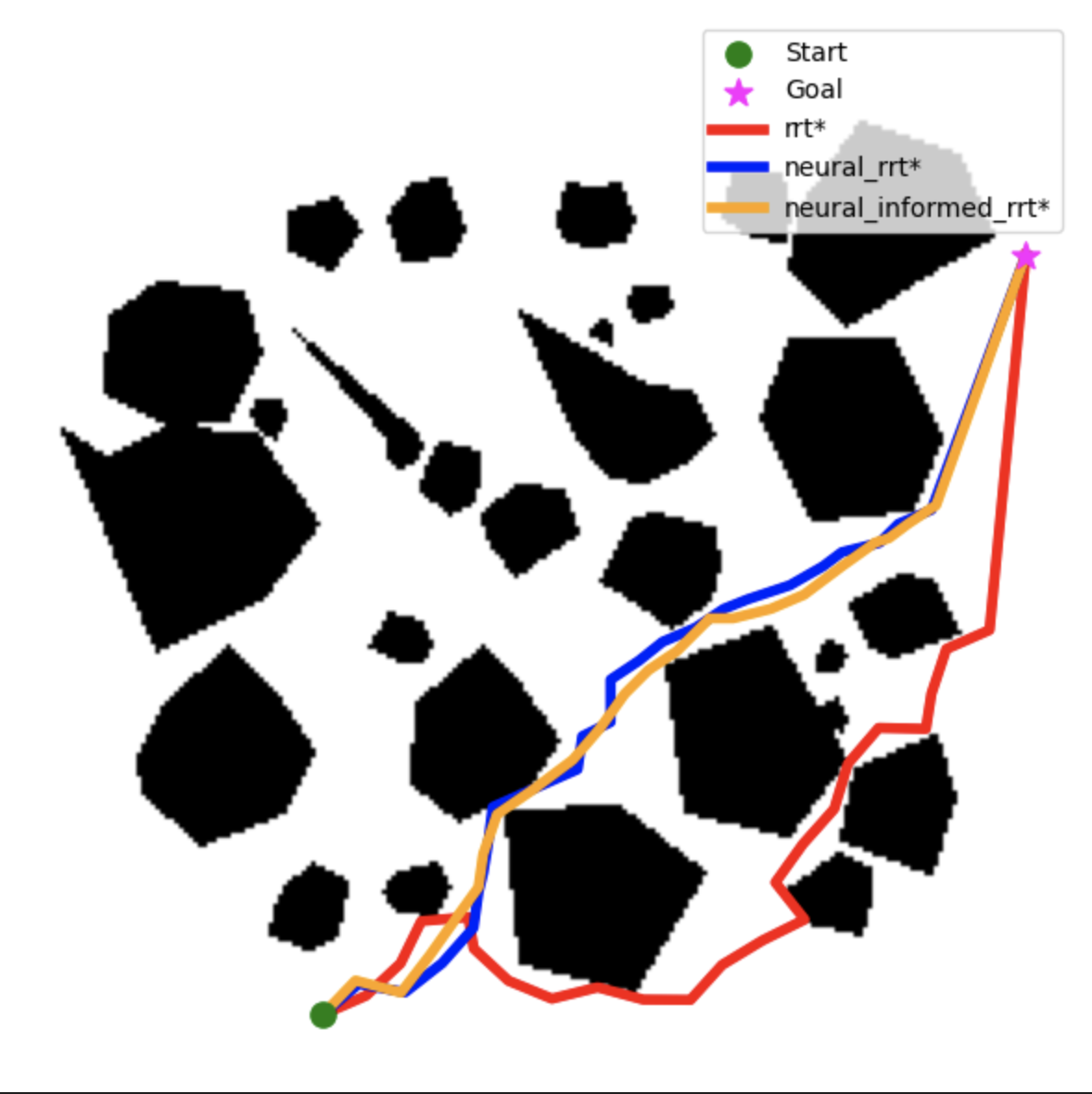}
        \caption{Medium-density maps}
        \label{fig:medium_comparison}
    \end{subfigure}
    \hfill
    \begin{subfigure}[b]{0.7\linewidth}
        \centering
        \includegraphics[width=\linewidth]{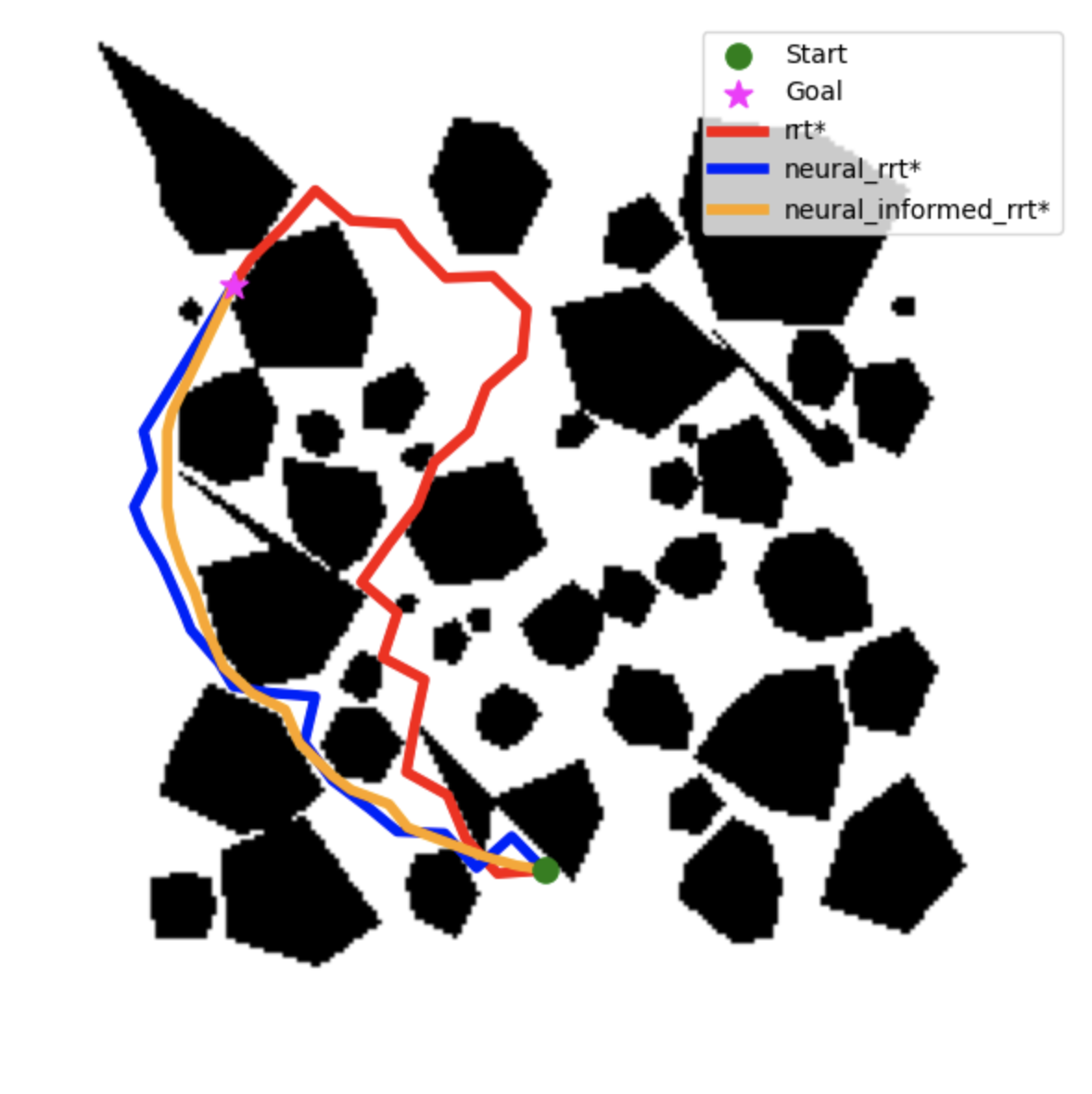}
        \caption{Dense maps}
        \label{fig:dense_comparison}
    \end{subfigure}
    \caption{Comparison of RRT*, Neural RRT*, and Neural Informed RRT* paths across different map densities. }
    \label{fig:all_maps_comparison}
\end{figure}

Quantitative results for the three planners across sparse, medium, and dense environments are summarized in Tables~\ref{tab:sparse_table_short}–\ref{tab:medium_table} and visualized in Fig.~\ref{fig:all_maps_comparison}

\begin{table}[htbp]
\centering
\caption{Performance of RRT*, Neural RRT*, and Neural Informed RRT* on sparse maps.}
\label{tab:sparse_table_short}
\setlength{\tabcolsep}{2pt}
\renewcommand{\arraystretch}{1.1}
\footnotesize
\begin{tabular}{llcccc}
\toprule
Map & Algorithm & Path Length & Time & Smoothness & Success  \\
\midrule
1 & RRT* & 206.17$\pm$11.95 & \textbf{0.34}$\pm$0.02 & 8.26$\pm$3.14 & 100.0 \\
  & Neural RRT* & 182.61$\pm$2.82 & 0.64$\pm$0.06 & 6.16$\pm$1.54 & 100.0 \\
  & Neural Inf~RRT* & \textbf{178.50}$\pm$0.75 & 0.59$\pm$0.03 & \textbf{3.71}$\pm$0.62 & 100.0 \\
\midrule
2 & RRT* & 130.04$\pm$6.75 & \textbf{0.35}$\pm$0.01 & 2.83$\pm$1.10 & 100.0 \\
  & Neural RRT* & 117.88$\pm$1.15 & 0.71$\pm$0.11 & 2.22$\pm$0.59 & 100.0 \\
  & Neural Inf~RRT* & \textbf{116.56}$\pm$0.60 & 0.57$\pm$0.04 & \textbf{1.61}$\pm$0.28 & 100.0 \\
\midrule
3 & RRT* & 295.18$\pm$20.08 & \textbf{0.30}$\pm$0.01 & 17.94$\pm$2.87 & 100.0 \\
  & Neural RRT* & 265.49$\pm$11.76 & 0.55$\pm$0.05 & 13.78$\pm$2.08 & 100.0 \\
  & Neural Inf~RRT* & \textbf{251.57}$\pm$9.99 & 0.51$\pm$0.03 & \textbf{9.84}$\pm$2.36 & 100.0 \\
\midrule
4 & RRT* & 219.68$\pm$20.59 & \textbf{0.33}$\pm$0.02 & 12.38$\pm$2.16 & 100.0 \\
  & Neural RRT* & 171.15$\pm$5.28 & 0.57$\pm$0.06 & 6.89$\pm$2.69 & 100.0 \\
  & Neural Inf~RRT* & \textbf{165.46}$\pm$2.28 & 0.55$\pm$0.03 & \textbf{4.37}$\pm$1.36 & 100.0 \\
\midrule
5 & RRT* & 176.30$\pm$2.11 & \textbf{0.36}$\pm$0.02 & 1.76$\pm$0.81 & 100.0 \\
  & Neural RRT* & 172.41$\pm$0.40 & 0.66$\pm$0.06 & 1.22$\pm$0.38 & 100.0 \\
  & Neural Inf~RRT* & \textbf{171.77}$\pm$0.13 & 0.59$\pm$0.01 & \textbf{1.13}$\pm$0.24 & 100.0 \\
\bottomrule
\end{tabular}
\end{table}

\begin{table}[htbp]
\centering
\caption{Performance of RRT*, Neural RRT*, and Neural Informed RRT* on medium-density maps.}
\label{tab:medium_table}
\setlength{\tabcolsep}{2pt}
\footnotesize
\renewcommand{\arraystretch}{1.1}
\begin{tabular}{llcccc}
\toprule
Map  & Algorithm & Path Length & Time & Smoothness & Success \\
\midrule
1 & RRT* & 249.29$\pm$8.76 & \textbf{0.29}$\pm$0.04 & \textbf{5.30}$\pm$1.63 & 100.0 \\
  & Neural RRT* & 230.05$\pm$13.92 & 0.39$\pm$0.09 & 6.96$\pm$1.55 & 100.0 \\
  & Neural Inf~RRT* & \textbf{216.98}$\pm$6.16 & 0.37$\pm$0.16 & 7.88$\pm$2.31 & 100.0 \\
\midrule
2 & RRT* & 134.01$\pm$17.57 & \textbf{0.30}$\pm$0.03 & 5.35$\pm$1.97 & 100.0 \\
  & Neural RRT* & 115.32$\pm$1.69 & 0.57$\pm$0.07 & 2.38$\pm$0.89 & 100.0 \\
  & Neural Inf~RRT* & \textbf{113.31}$\pm$0.18 & 0.59$\pm$0.03 & \textbf{1.27}$\pm$0.26 & 100.0 \\
\midrule
3 & RRT* & 107.21$\pm$0.25 & \textbf{0.33}$\pm$0.01 & 0.19$\pm$0.18 & 100.0 \\
  & Neural RRT* & 107.02$\pm$0.00 & 0.70$\pm$0.13 & 0.02$\pm$0.01 & 100.0 \\
  & Neural Inf~RRT* & \textbf{107.02}$\pm$0.00 & 0.63$\pm$0.01 & \textbf{0.01}$\pm$0.01 & 100.0 \\
\midrule
4 & RRT* & 146.16$\pm$16.29 & \textbf{0.31}$\pm$0.03 & 5.25$\pm$1.98 & 100.0 \\
  & Neural RRT* & 121.59$\pm$0.71 & 0.70$\pm$0.14 & 2.09$\pm$0.37 & 100.0 \\
  & Neural Inf~RRT* & \textbf{120.92}$\pm$0.31 & 0.65$\pm$0.03 & \textbf{1.49}$\pm$0.31 & 100.0 \\
\midrule
5 & RRT* & 143.32$\pm$11.96 & \textbf{0.33}$\pm$0.03 & 5.79$\pm$2.42 & 100.0 \\
  & Neural RRT* & 120.36$\pm$1.07 & 0.65$\pm$0.05 & 2.40$\pm$0.83 & 100.0 \\
  & Neural Inf~RRT* & \textbf{118.70}$\pm$0.16 & 0.59$\pm$0.04 & \textbf{1.33}$\pm$0.29 & 100.0 \\
\bottomrule
\end{tabular}
\end{table}

\begin{table}[htbp]
\centering
\caption{Performance of RRT*, Neural RRT*, and Neural Informed RRT* on dense maps }
\label{tab:medium_table}
\setlength{\tabcolsep}{2pt}
\footnotesize
\renewcommand{\arraystretch}{1.1}
\begin{tabular}{llcccc}
\toprule
Map & Algorithm & Path Length & Time & Smoothness & Success \\
\midrule
1 & RRT* & 240.83 ± 31.45 & \textbf{0.22} ± 0.04 & 13.46 ± 3.10 & 100.0 \\
 & Neural RRT* & 224.30 ± 27.42 & 0.38 ± 0.05 & 12.09 ± 3.67 & 100.0 \\
 & Neural Inf~RRT* & \textbf{225.39}± 29.99 & 0.28 ± 0.07 & \textbf{13.87} ± 4.15 & 100.0 \\
\midrule
 & RRT* & 180.64 ± 19.28 & \textbf{0.27} ± 0.02 & 10.24 ± 2.90 & 100.0 \\
2 & Neural RRT* & 158.37 ± 5.00 & 0.52 ± 0.08 & 6.52 ± 1.76 & 100.0 \\
 & Neural Inf~RRT* & \textbf{150.59} ± 1.75 & 0.53 ± 0.02 & \textbf{4.06} ± 0.82 & 100.0 \\
\midrule
3 & RRT* & 242.32 ± 26.15 & \textbf{0.23} ± 0.05 & 15.17 ± 2.83 & 100.0 \\
 & Neural RRT* & 190.87 ± 12.18 & 0.45 ± 0.05 & 12.25 ± 2.76 & 100.0 \\
 & Neural Inf~RRT* & \textbf{177.27} ± 12.49 & 0.48 ± 0.13 & \textbf{9.62} ± 3.24 & 100.0 \\
\midrule
4 & RRT* & 266.17 ± 13.54 & \textbf{0.29} ± 0.04 & 10.50 ± 2.44 & 100.0 \\
 & Neural RRT* & 258.50 ± 15.09 & 0.46 ± 0.07 & 11.07 ± 2.39 & 100.0 \\
 & Neural Inf~RRT* & \textbf{229.98} ± 6.41 & 0.35 ± 0.12 & \textbf{6.37} ± 1.58 & 90.0 \\
\midrule
5 & RRT* & 137.17 ± 0.59 & \textbf{0.28} ± 0.02 & 0.32 ± 0.29 & 100.0 \\
 & Neural RRT* & 136.69 ± 0.17 & 0.60 ± 0.10 & 0.11 ± 0.11 & 100.0 \\
 & Neural Inf~RRT* & \textbf{136.62} ± 0.00 & 0.60 ± 0.01 & \textbf{0.01} ± 0.01 & 100.0 \\
\bottomrule
\end{tabular}
\end{table}

\textbf{Path length and smoothness:} Neural Informed RRT* consistently achieves the shortest and smoothest paths. For example, in dense Map 2, the path length is reduced from $242.32 \pm 26.15$ (RRT*) to $177.27 \pm 12.49$, while smoothness improves from $15.17 \pm 2.83$ to $9.62 \pm 3.24$ (Table~\ref{tab:medium_table}). Neural RRT* also improves path quality, but to a lesser extent. These trends are clearly visible in Fig.~\ref{fig:dense_comparison}, where the paths generated by Neural Informed RRT* are smoother and closer to the optimal trajectory.

\textbf{Planning time:} Planning time remains a key differentiator among the algorithms. Standard RRT* is generally faster in sparse maps, e.g., $0.34 \pm 0.02$ s in sparse Map 1 (Table~\ref{tab:sparse_table_short}). However, as obstacle density increases, the computational overhead of neural guidance persists but is outweighed by the significant improvements in path quality. In dense Map 4, for example, RRT* requires $0.29 \pm 0.04$ s versus $0.35 \pm 0.12$ s for Neural Informed RRT*. This difference confirms that the neural network's inference time adds a measurable, consistent cost. Nevertheless, this marginal increase in computation time yields paths that are up to 75\% smoother and shorter. Therefore, the higher planning time is an acceptable cost for the enhanced trajectory efficiency and reliability provided by the neural-guided planners.

\textbf{Success rate:} All three planners maintain nearly 100\% success in sparse and medium-density maps. In dense maps, Neural Informed RRT* occasionally drops to 90\% (dense Map 3), likely due to aggressive pruning of the search space, but still outperforms the others in path efficiency and smoothness.

Overall, the Neural Informed RRT* planner provides the best comparison metrics between path quality and computation time, especially in complex environments, making it the most suitable choice for robot navigation. Neural RRT* offers moderate improvement over RRT*, while RRT* remains competitive only in simple, sparse environments.

\subsection{Discussion:}
The superior performance of the neural planners arises from their learned sampling distributions, which guide exploration toward promising regions of the configuration space. Neural RRT* directs sampling using learned spatial patterns, reducing redundant exploration, while Neural Informed RRT* further enhances this process through adaptive reweighting that balances exploration and exploitation. This integration of learning-based inference with classical planning leads to faster convergence and higher-quality paths, especially in complex environments.

\section{Conclusion}
This study evaluated classical RRT*, Neural RRT*, and Neural Informed RRT* across sparse, medium, and dense map environments. The results show that Neural Informed RRT* consistently achieves shorter path lengths and smoother trajectories while maintaining high success rates, even in complex and cluttered maps. Neural RRT* also improves path efficiency compared to classical RRT*, though with slightly higher computation time. These findings highlight the advantages of integrating AI into robotic navigation, enabling faster, more reliable, and adaptive path planning. Future work will extend these approaches to dynamic, three-dimensional environments and real-world robotic applications.

\section*{Acknowledgment}
This work was supported by the research project "Modeling and Control of Aerial Manipulators" N° C00L07UN310220230004.

\end{document}